\documentclass{article}

\usepackage[left=3cm,right=3cm,top=0cm,bottom=2cm]{geometry} 
\usepackage{amsmath} 
\usepackage{amssymb}
\usepackage{mathrsfs}
\usepackage{graphicx}
\usepackage{subfigure}
\usepackage{mathtools}
\usepackage{cool}
\usepackage{relsize}
\usepackage{float}
\usepackage{bigints}

\begin{document}

\title{ {\bf Entropy Based Cartoon Texture Separation}}
\date{}
\author{KUTLU EMRE YILMAZ (keylmz @ gmail )}
\maketitle

\subsection*{Abstract}

Separating an image into cartoon and texture components comes useful in image processing applications such as image compression, image segmentation, image inpainting. Yves Meyer's influential cartoon texture decomposition model involves deriving an energy functional by choosing appropriate spaces and functionals.  Minimizers of the derived energy functional are cartoon and texture components of an image. In this study, cartoon part of an image is separated, by reconstructing it from  pixels of multi scale Total-Variation filtered versions of the original image which is sought to be decomposed into cartoon and texture parts. An information theoretic pixel by pixel selection criteria is employed to choose the contributing pixels and their scales.

%
%
%
%
%

\subsection*{Introduction}
An image, $f $, is defined as a function, \[ f : (x,y) \in \Omega \rightarrow \mathbb{R},   \] where $ f  $ maps a pixel location   
 $  (x,y) \in \Omega $  to an intensity value $ I \in \mathbb{R}  $, and $\Omega$ is the set of pixel locations. Cartoon-Texture separation problem is about decomposing $f$ into two functions, $u$ and $v$, where function $u$ is the cartoon part and function $v$ is the texture part of an image $f$, such that $f = u +v$. 

Total Variation noise removal algorithm [1] was introduced by Rudin, Osher and Fatemi (ROF). Instead of using $L_2$ norm in image processing related estimation problems, they used Total Variation(TV) norm, $L_1$ norm of gradient, which is the true norm for image processing related estimation problems. They mention that the space of functions with bounded TV norm is the space of functions of bounded variations. ROF model computes,  minimizer, $u$, of the following functional, $E(u)$ , where\\

$E(u) =  \bigints_{\Omega} | \nabla u |  + \dfrac{\lambda}{2} \bigints_{\Omega} (f - u)^2 dx \ dy$

\textbf{In this paper, the term scale in used to denote different values parameter $\lambda$ may get. }

While space of functions of bounded variations represent edges and piecewise smooth parts of an image, space of oscillatory functions represent texture and noise. ROF model is useful for computing an approximation to cartoon part $u$. Yves Meyer [2] devised a new functional, $E(u,v)$ and introduced a second minimizer, $v$ representing texture and noise component of an image $f$. In Buades et al. Meyer's approach is defined as , 

\newpage
\vspace*{50px}

 \[ E(u,v) = \adjustlimits \inf_{(u,v) \in X_1, X_2 } \{f = u+v : F_1(u) + F_2(v) \}  \]

where $F_1, F_2 \geq 0 $ are functionals, and $ X_1, X_2 \geq 0 $ are spaces of functions. Functional $F_1(u)$ penalizes texture component and functional $F_2(v)$ penalizes cartoon component, so the aim is minimizing with respect to both variables, $u$ and $v$ to achieve a perfect cartoon texture separation. Many related study focus on the effect of choosing different function spaces  and functionals for the $v$ component. e.g. spaces G, F, E of various generalized functions. Yin et al. [3] gives a detailed comparison of these function space approaches and their performance.  Although, choices of many function spaces and functionals are possible for texture component, $v$, cartoon component, $u$, is always modelled by the space of $L_1$ norm of the gradient.

ROF functional is convex, so solution of the following Euler-Lagrange equation,

 \[  u - f - div  \left(	\frac{\nabla u}{|\nabla u|}	\right) = 0 \]

is the minimizer, u , of the ROF energy $E(u)$. To compute, $u$, gradient descent iterations on the direction  $ - \dfrac{dE(u)}{du}$ is applied which leads to the following PDE,

\[      \dfrac{ \partial u }{ \partial t }    = \lambda \left( f - u \right) +  div  \left( \frac{\nabla u}{|\nabla u|}  \right)       \]

Main problem on implementation of a numerical scheme for gradient descent iterations is when gradient component,$|\nabla u|$,  is zero. A simple solution is, replacing                    $|\nabla u|$ term with $\sqrt{\epsilon^2 + {\nabla u}^2}$ by adding a very small $\epsilon$ and carrying on time dependent step by step iterations. However due to CFL conditions and small $\epsilon$ value these iterations are very slow to converge. 

Many other optimization methods, such as primal-dual methods, gradient projection methods, graph cut algorithms were proposed to minimize ROF energy. A detailed review about efficient approaches for computing ROF minimizers is given in Zhu et al. [4]. The most notable one is Chan, Golub, Mulet (CGM)Algorithm [5] which I preferred to use in my experiments. URL of a publicly available implementation is given in [6]. The reason for preference was its fast convergence rate.

Entropy, $H(\mathbf X)$, of a random variable $\mathbf{X}$ is defined as,

\[    \displaystyle - \sum \limits_{i}   p(x_i)  \log p(x_i)    \]

where $x_i$ are the values random variable $\mathbf{X}$ maps to.

According to this definition, uniform distribution must have the maximum entropy i.e. highest uncertainty among all continuous probability distributions since each value of uniform random variable (RV) has equal probability of occurring. 

\newpage
\vspace*{50px}

Definition of a texture is not clear, however it can be interpreted as structured image segments following a regular or stochastic pattern. Detecting this pattern is subject to one's perceptual selectivity. In this study textures can be classified as regular, near-regular, irregular, near-stochastic and stochastic.

\begin{figure}[H]
\centering
\includegraphics[scale=0.40]{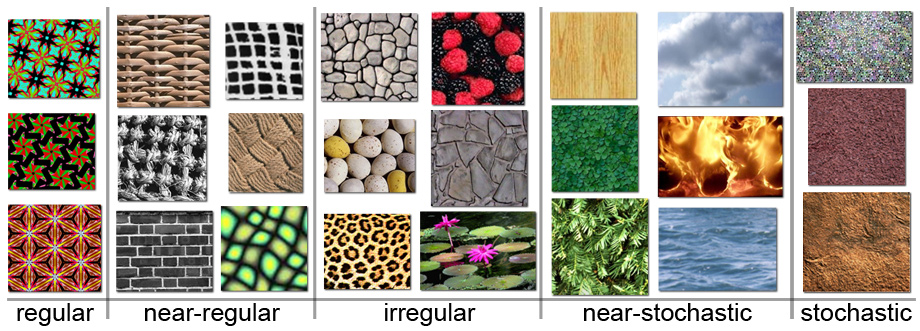}
\caption{Classification of textures. Figure is from  "Near-regular Texture Analysis and Manipulation." Yanxi Liu, Wen-Chieh Lin, and James Hays. SIGGRAPH 2004}
\label{fig:figure}
\end{figure}

%
%
%
%
%

\subsection*{Textures and Entropy}
In this study, my focus is on real world images with textures on them. My hypothesis stems from the fact that, intensity of pixels in a gray scale image texture have a more uniform probability distribution due to same texture elements repeating regularly in a texture, especially this is valid for regular and near-regular textures. 

A uniform distribution leads to higher entropy and increased uncertainty. I propose that, investigating the entropy of a local $k \times k$ block around each pixel in an image, gives us a clue about either that pixel is a texture element or not. After deciding whether a pixel is a texture element or not, we choose replace it with the appropriate cartoonized pixel from the corresponding ROF scales.

Definition: Let $f$ be a real world image with a pixel $p \in f$, then if $p$ belongs to texture component, we denote it as $p_t$, and otherwise we denote is as $p_c$.

Proposition: In a real world image $f$, entropy of a local $ k  \times k $ block formed around a texture pixel, $p_t$, where $p_t$ is at the center of the block, is always greater than the entropy of a local $k \times k$ block formed around a cartoon pixel, $p_c$ where $p_c$ is also at the center of its block.

An intuitive justification of this proposition is due to a more  uniform distribution $k \times k$ blocks around a texture pixel $p_t$ has. This leads to higher entropy values. The entropy estimation procedure is based on only $k \times k$ sample. Also, all pixels in a block assumed to be independently and identically distributed. Probability of a pixel with intensity value, $I$, occurring in a block, is the fraction of the number of occurrences of $I$ to number of all pixels in that block, namely $k \times k$. In natural images, intensity values of neighbour pixels are highly correlated. This is a drawback in our estimation procedure.

Next, I give some experiments with the entropy  of real world images.

%
%
%
%
%

\newpage
\vspace*{50px}

\subsection*{Structure type images with texture.}
Textures on building images are perfectly separated by entropy calculation. Also, entropy difference between sea and building walls is obvious. Since sea has a regular color structure, its entropy is lower than the entropy of walls.

\begin{figure}[H]
\centering
\subfigure[]{%
\includegraphics[scale=0.15]{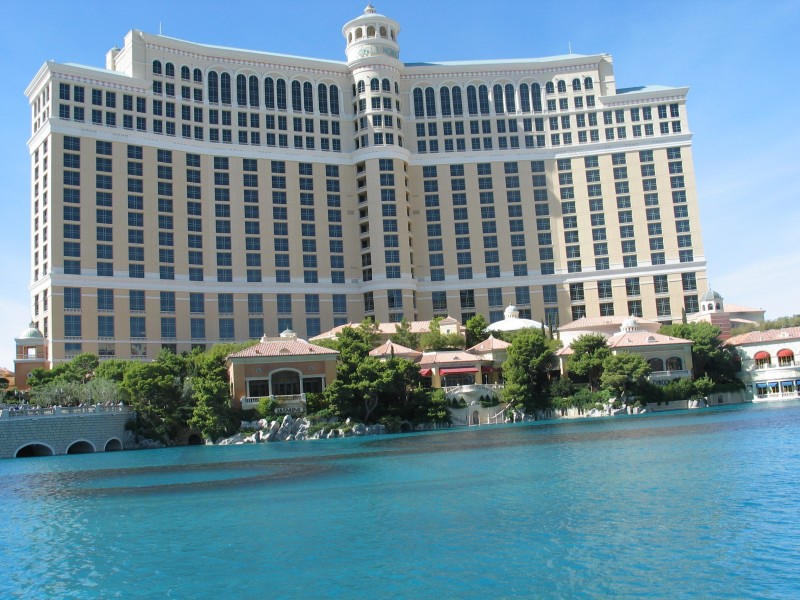}
\label{fig:subfigure1}}
\quad
\subfigure[]{%
\includegraphics[scale=0.15]{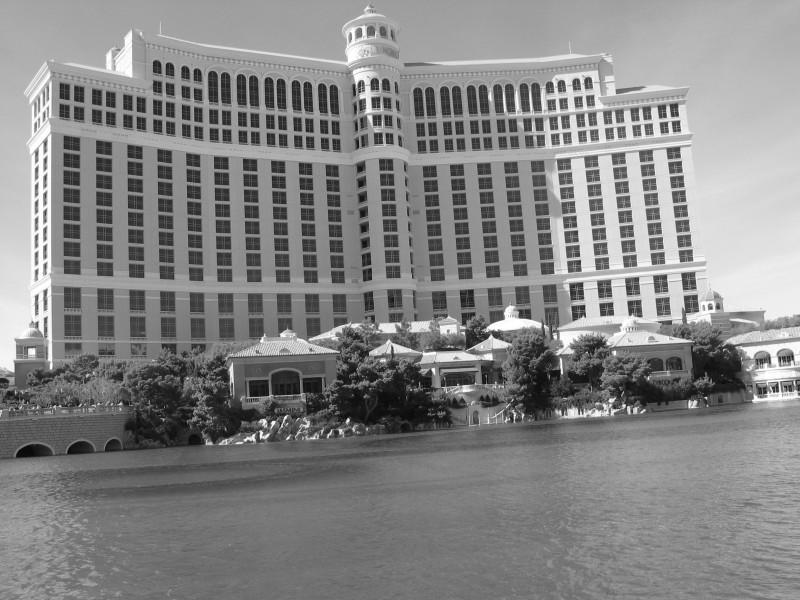}
\label{fig:subfigure3}}
\quad
\subfigure[]{%
\includegraphics[scale=0.14]{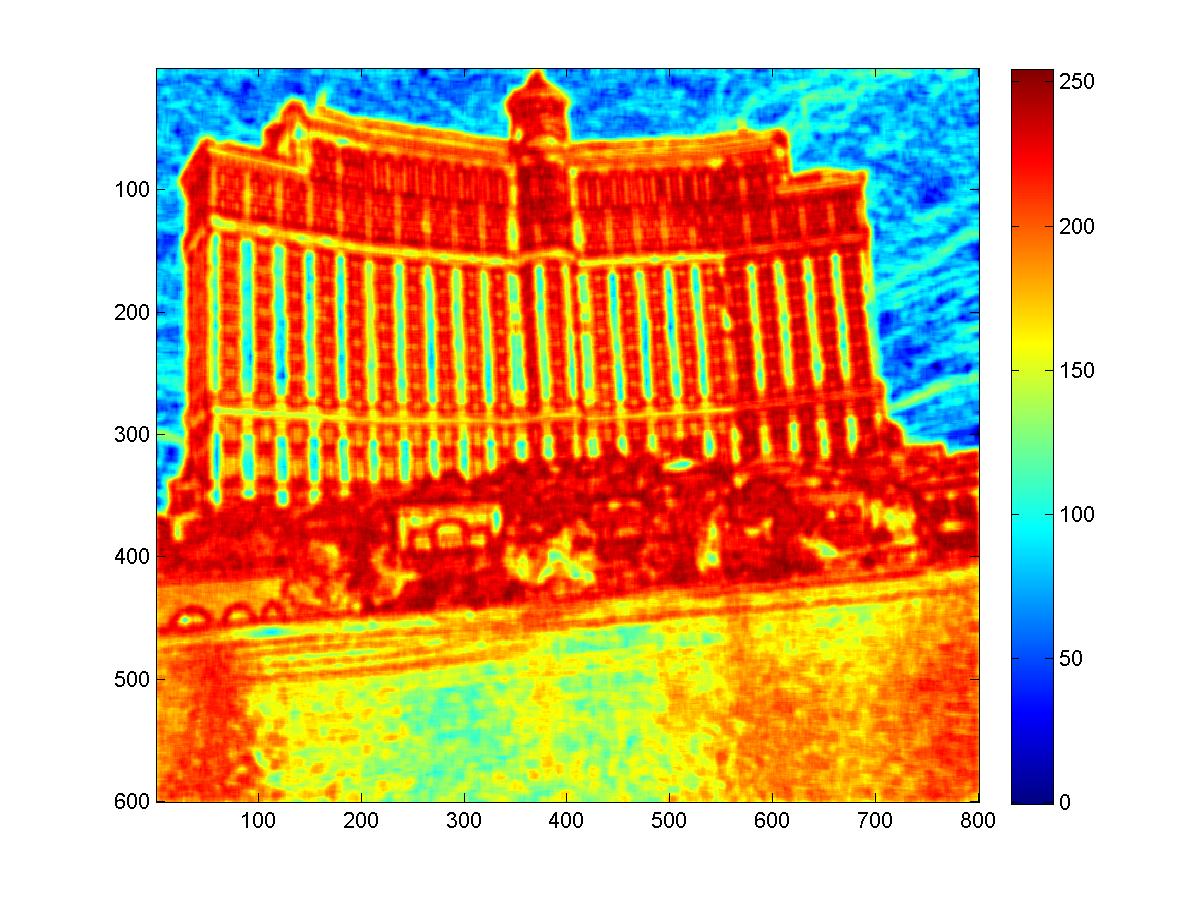}
\label{fig:subfigure4}}
\caption{Example image of a building type.}
\label{fig:figure}
\end{figure}

An example of a structure type image, its gray scale version and entropy calculated on its gray scale version. Entropy calculation is made by calculating the entropy of each $9 \times 9$ pixel block around every pixel. Resulting entropy values are mapped into range $[0, 255]$. Jet colormap in MATLAB is used to plot entropy values.   Images are from  PSU Texture Database [7].

%
%
%
%
%

\subsection*{Cloth type images with texture.}

Neighbour texture regions (Red and Blue rectangles) in sweater image has nearly same entropy values. However, a careful eye can distinguish the slight difference between them. 

\newpage
\vspace*{50px}

\begin{figure}[H]
\centering
\subfigure[]{%
\includegraphics[scale=0.15]{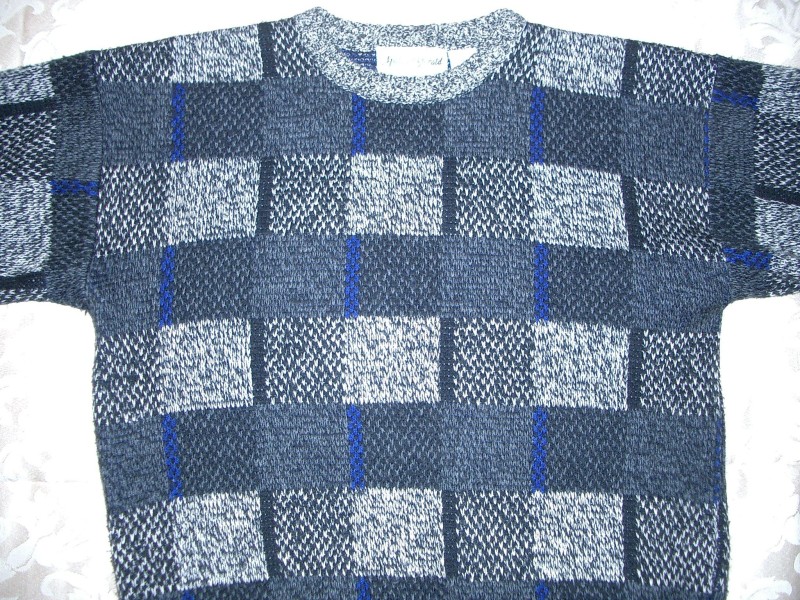}
\label{fig:subfigure1}}
\quad
\subfigure[]{%
\includegraphics[scale=0.20]{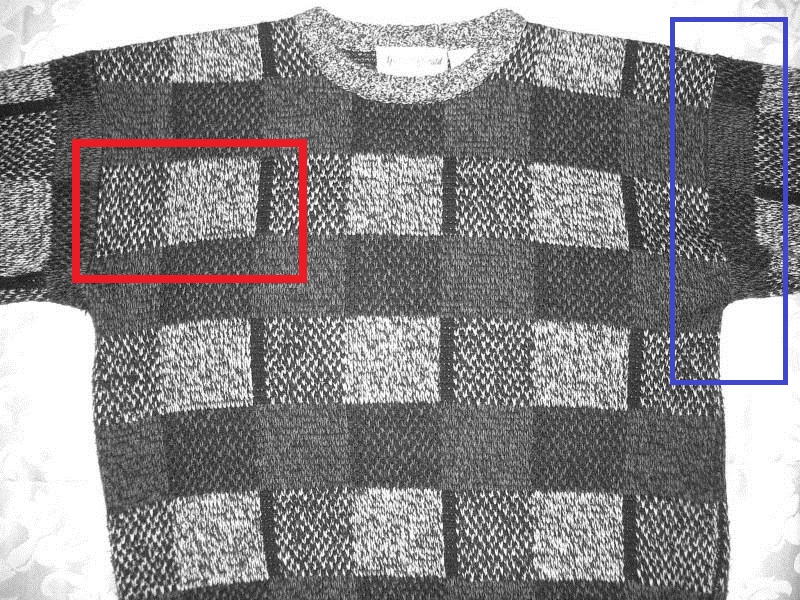}
\label{fig:subfigure3}}
\quad
\subfigure[]{%
\includegraphics[scale=0.14]{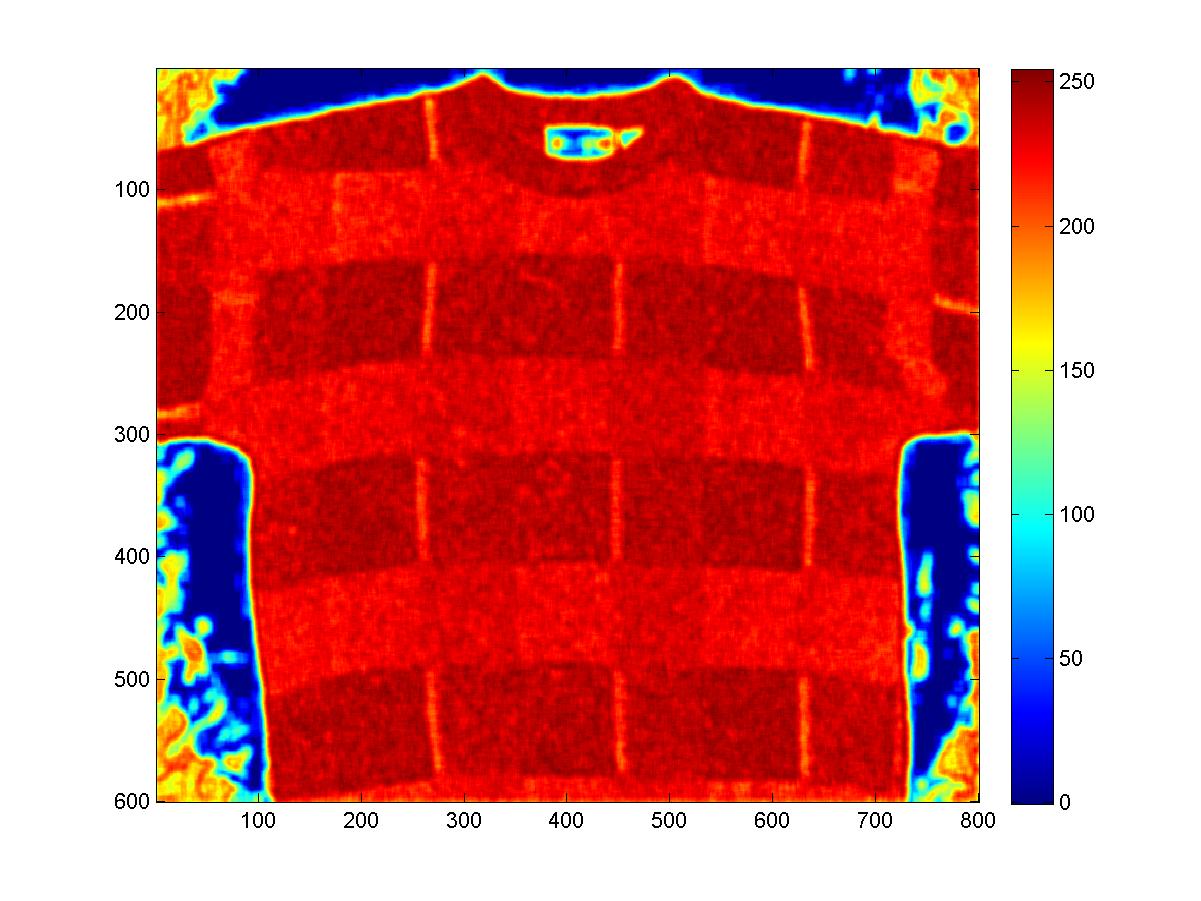}
\label{fig:subfigure4}}
\caption{Example image of cloth  type.}
\label{fig:figure}
\end{figure}

An example of a cloth type image, its gray scale version and entropy calculated on its gray scale version. Entropy calculation is made by calculating the entropy of each $9 \times 9$ pixel block around every pixel. Resulting entropy values are mapped into range $[0, 255]$. Jet colormap in MATLAB is used to plot entropy values.   Images are from  PSU Texture Database[7].

%
%
%
%
%

\subsection*{Screen type images with texture.}

Most interesting result here is that, how this method generalizes not only to textures which are local regions in an image, but also to textures which has a global structure in an image.

 An example of a screen type image, its gray scale version and entropy calculated on its gray scale version. Entropy calculation is made by calculating the entropy of each $9 \times 9$ pixel block around every pixel. Resulting entropy values are mapped into range $[0, 255]$. Jet colormap in MATLAB is used to plot entropy values.  Images are from  PSU Texture Database[7].

 \newpage
\vspace*{50px}

\begin{figure}[H]
\centering
\subfigure[]{%
\includegraphics[scale=0.15]{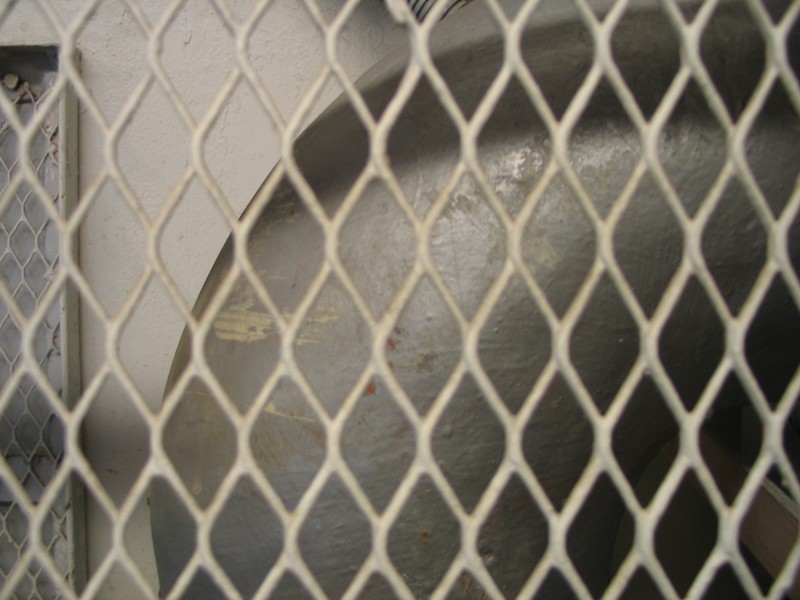}
\label{fig:subfigure1}}
\quad
\subfigure[]{%
\includegraphics[scale=0.15]{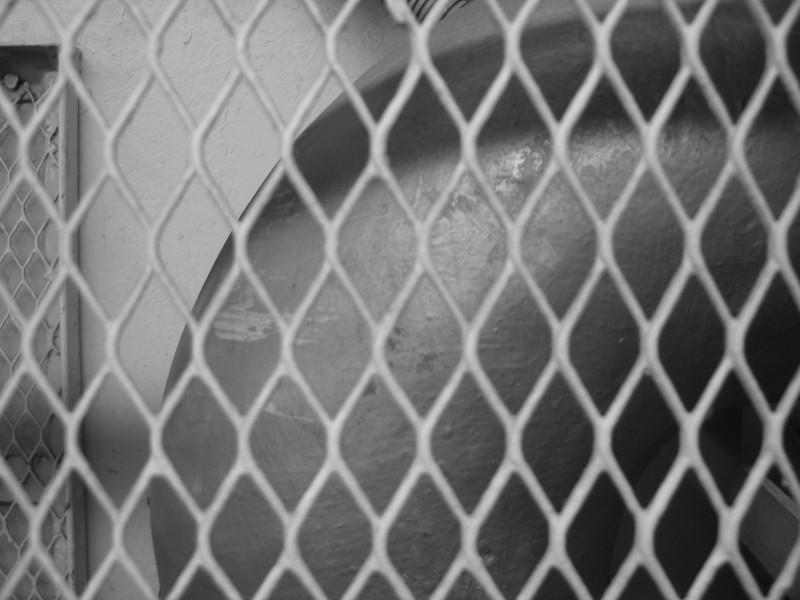}
\label{fig:subfigure3}}
\quad
\subfigure[]{%
\includegraphics[scale=0.14]{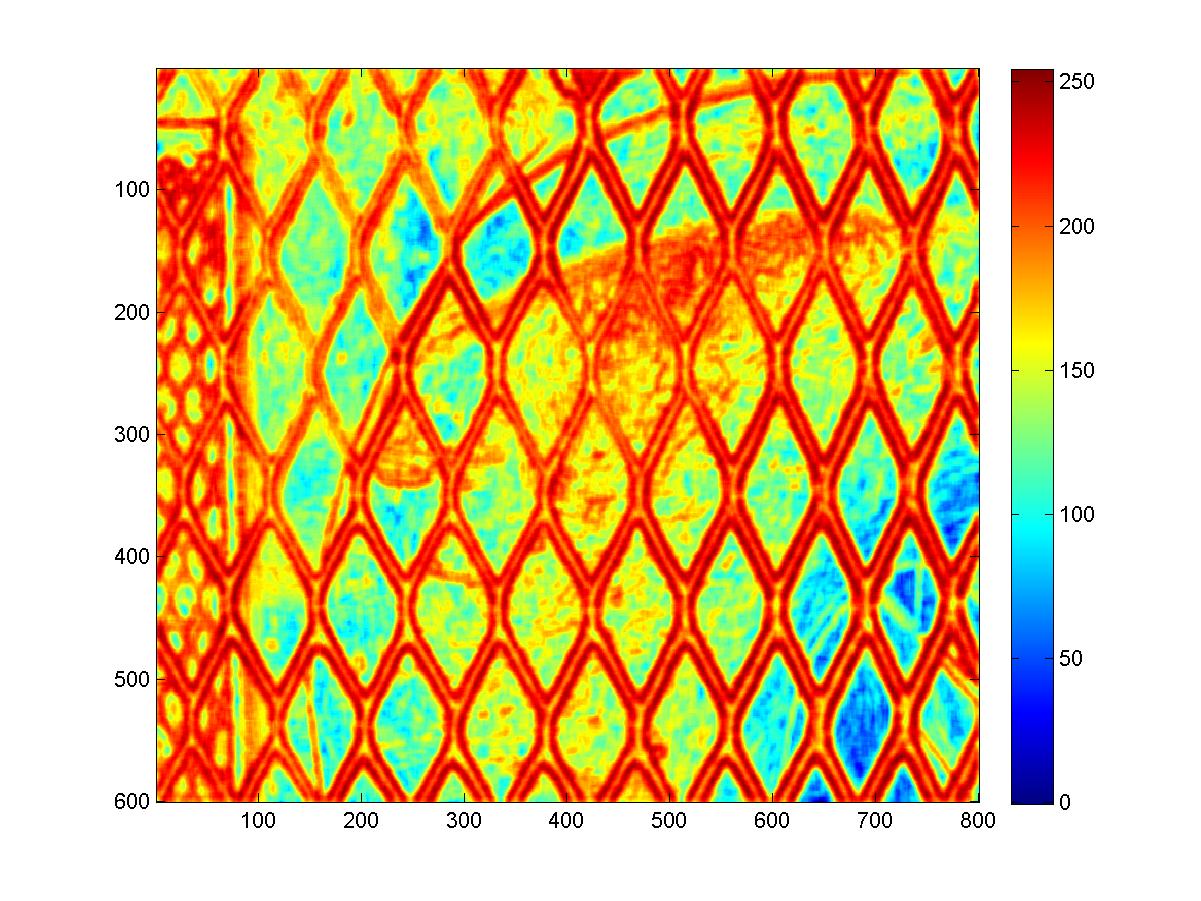}
\label{fig:subfigure4}}
\caption{Example image of screen  type.}
\label{fig:figure}
\end{figure}

%
%
%
%
%

\subsection*{Randomly chosen various images with texture.}

Notice  how entropy of a blackboard image is low (colored in blue) due to its constant color structure. Clothes of person  are in red, also, red yellow color distribution on man's sweater discriminates two different texture.

\begin{figure}[H]
\centering
\subfigure[]{%
\includegraphics[scale=0.15]{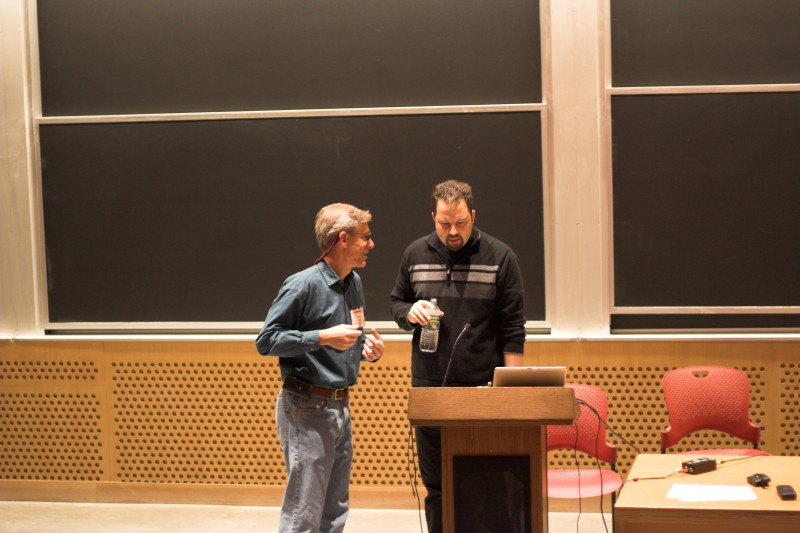}
\label{fig:subfigure1}}
\quad
\subfigure[]{%
\includegraphics[scale=0.15]{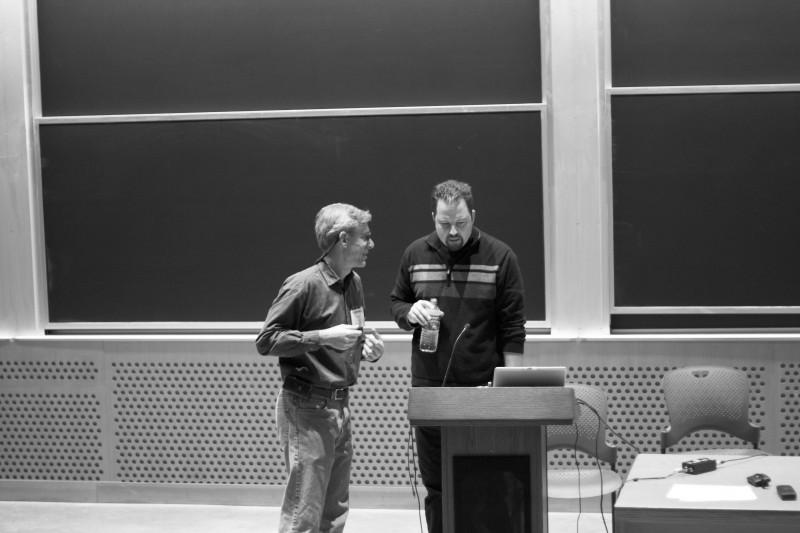}
\label{fig:subfigure3}}
\quad
\subfigure[]{%
\includegraphics[scale=0.14]{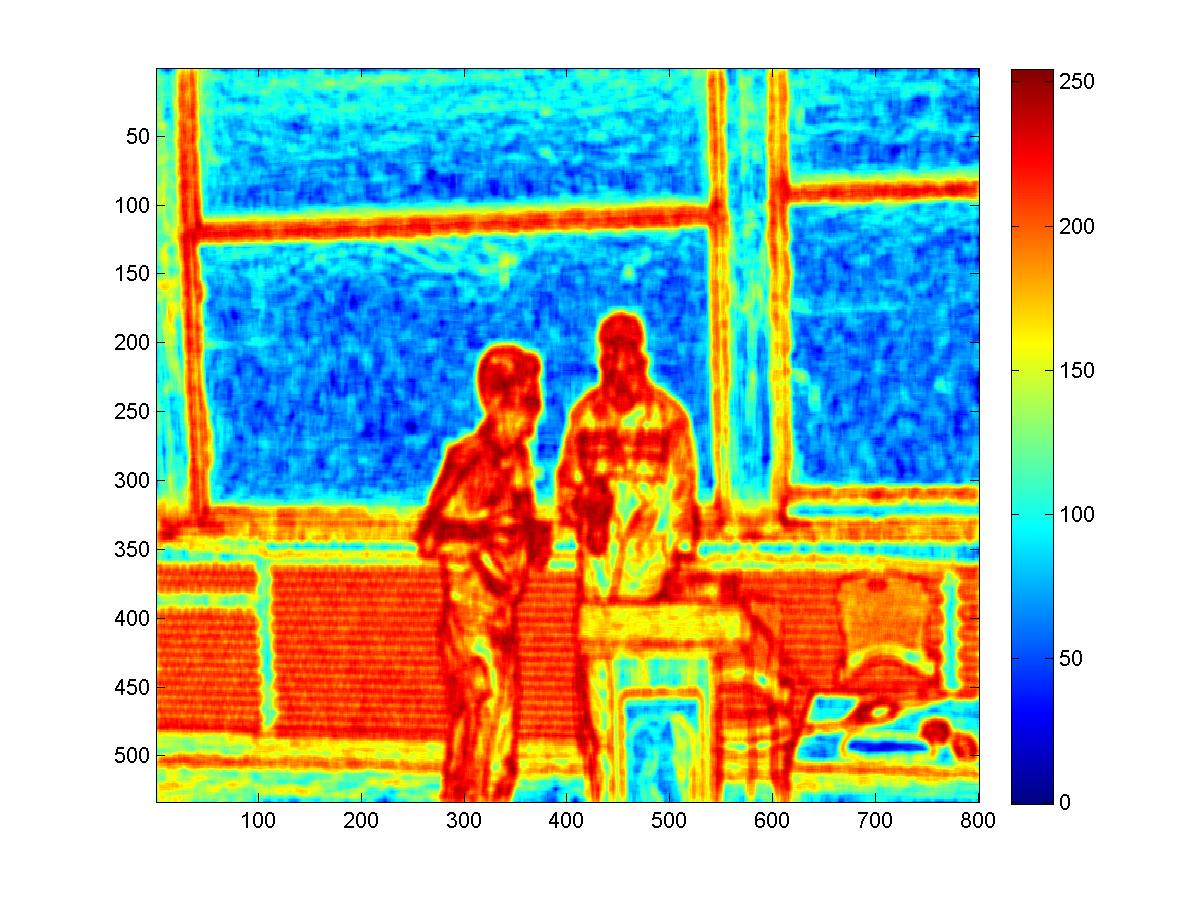}
\label{fig:subfigure4}}
\caption{Example of a randomly chosen image.}
\label{fig:figure}
\end{figure}

 \newpage
\vspace*{50px}
An example of a randomly selected  image, its gray scale version and entropy calculated on its gray scale version. Entropy calculation is made by calculating the entropy of each $9 \times 9$ pixel block around every pixel. Resulting entropy values are mapped into range $[0, 255]$. Jet colormap in MATLAB is used to plot entropy values.   Images are from  PSU Texture Database [7].

%
%
%
%
%

\subsection*{Classic test images.}

Barbara is a classic test image and a perfect example with its local textures. The floor, walls and chair has the lowest entropy with their piecewise constant color. The scarf and  trousers of woman, table cloth has the highest entropy value.
 
\begin{figure}[H]
\centering
\subfigure[]{%
\includegraphics[scale=0.15]{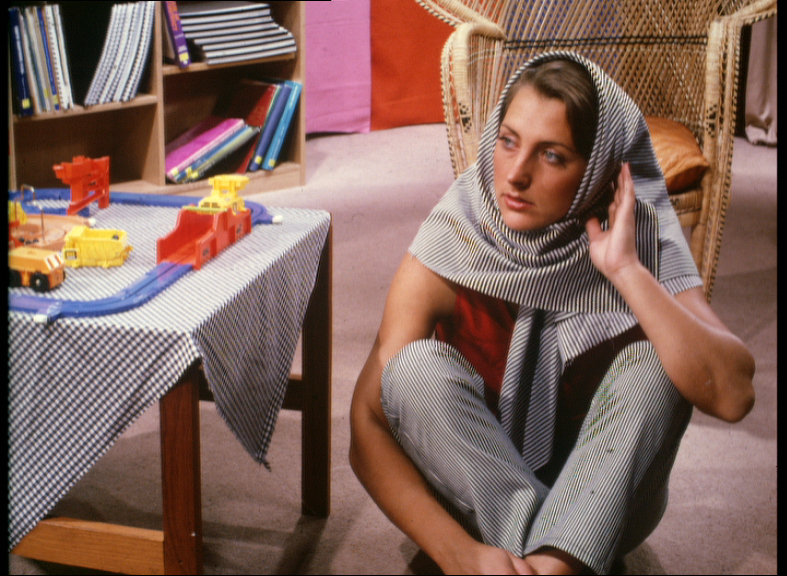}
\label{fig:subfigure1}}
\quad
\subfigure[]{%
\includegraphics[scale=0.15]{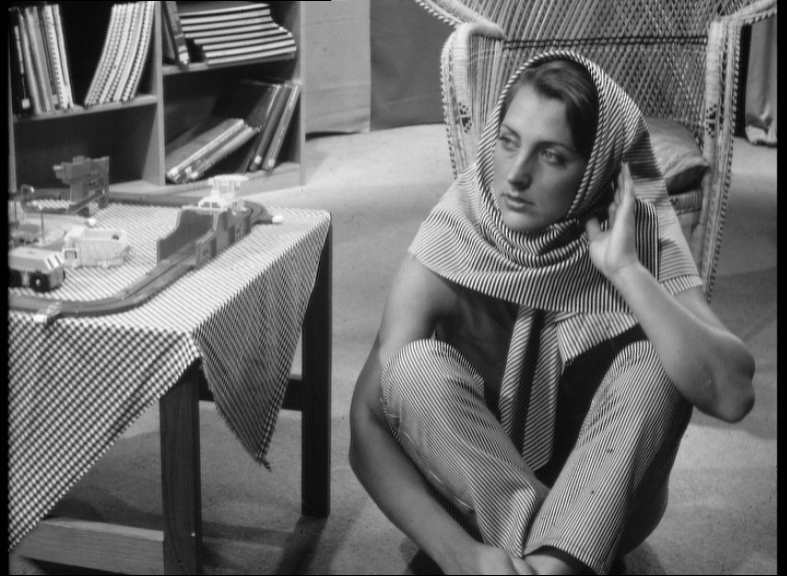}
\label{fig:subfigure3}}
\quad
\subfigure[]{%
\includegraphics[scale=0.5]{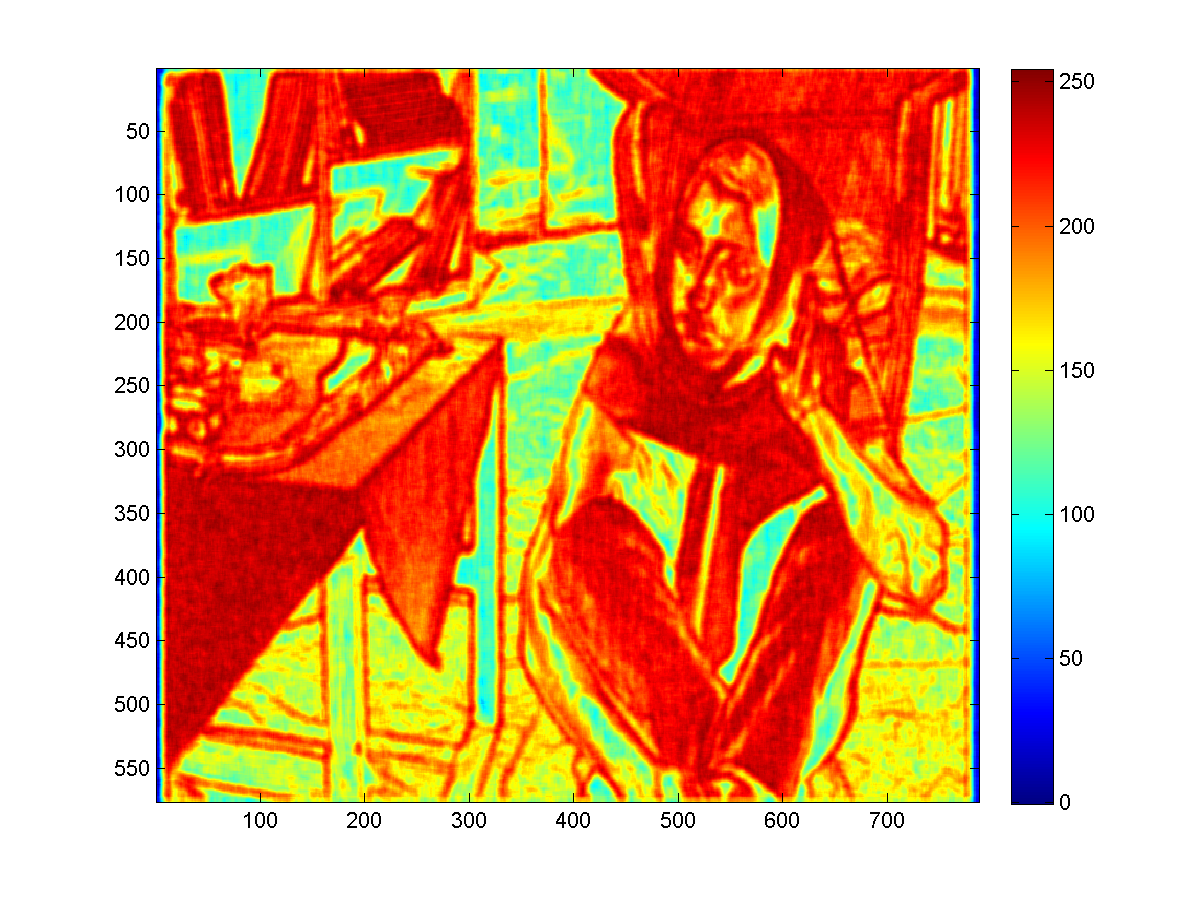}
\label{fig:subfigure4}}
\caption{Classic test image Barbara .}
\label{fig:figure}
\end{figure}

An example of a classic test image, its gray scale version and entropy calculated on its gray scale version. Entropy calculation is made by calculating the entropy of each $9 \times 9$ pixel block around every pixel. Resulting entropy values are mapped into range $[0, 255]$. Jet colormap in MATLAB is used to plot entropy values.   Images are from  PSU Texture Database [7].

%
%
%
%
%

 \newpage
\vspace*{50px}

\subsection*{Multi scale ROF examples}

ROF model has a single parameter $\lambda$ for tuning its fidelity term. If $\lambda$ is zero, then we can minimize $E(u)$ by choosing $u$ as a constant colored image which leads to zero gradient and energy $E(u)=0$.  If lambda is close to infinity, then choosing minimizer $u$ as the original image $f$ still leads to minimized energy $E(u)=0$. Different lambda values give different cartoonized images. The most important thing is preserving edges that  exist on original image $f$, when TV Filtering is applied. Choosing very small $\lambda$ eliminates almost all textures on the image at the expense of blurring some important edges on original image $f$. These types of edges are usually the ones that can still be preserved by choosing a larger $\lambda$ value without comprising from expected cartoonization behaviour.

Notice how the edge behaviour of objects in red and green rectangular regions change with $ \lambda $ value. We lose edge structure of most objects very early. So,  pixels contributing to final cartoon image created by our algorithm must be selected carefully to avoid this problem. 

\begin{figure}[H]
\centering
\subfigure[Original Image]{%
\includegraphics[scale=0.4]{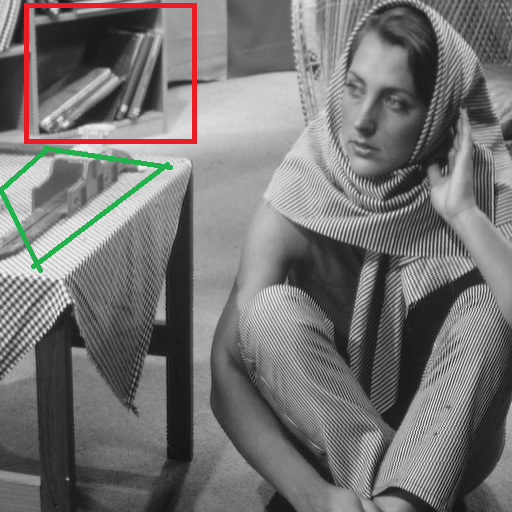}
\label{fig:subfigure1}}
\quad
\subfigure[$\lambda = 0.045$]{%
\includegraphics[scale=0.3]{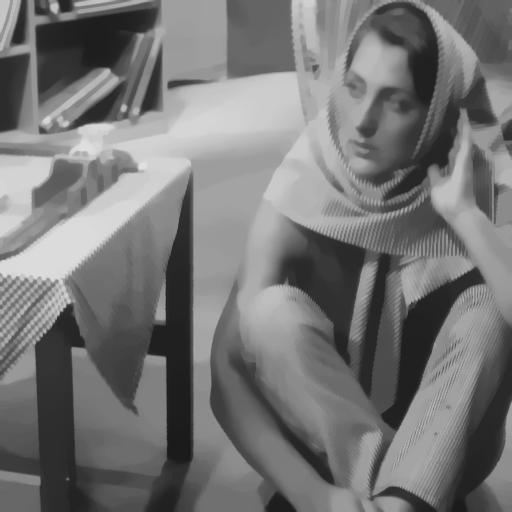}
\label{fig:subfigure2}}
\subfigure[$\lambda = 0.025$]{%
\includegraphics[scale=0.3]{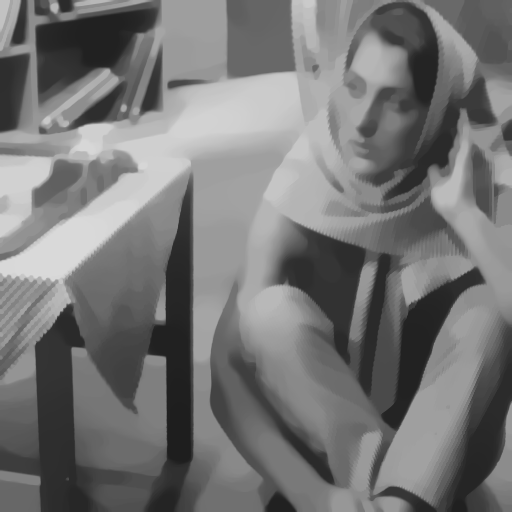}
\label{fig:subfigure3}}
\quad
\subfigure[$\lambda = 0.005$]{%
\includegraphics[scale=0.3]{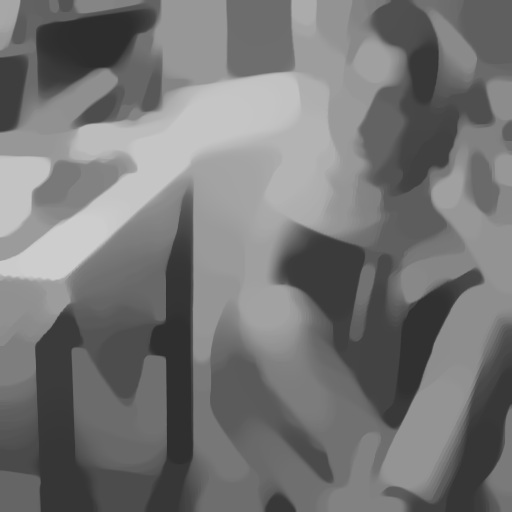}
\label{fig:subfigure4}}
\caption{ TV Filtering applied images with different  $\lambda$ values . }
\label{fig:figure}
\end{figure}

%
%
%
%
%

 \newpage
\vspace*{50px}

\subsection*{Algorithm.}

What inspired me the most was the article "Fast cartoon + texture image filters" of Buades et al. [8]. They use local total variation around each pixel at different Gaussian scales to decide whether that pixel is a texture element or not. If a pixel is a texture element, they replace it with a low pass filtered version of it.

In my approach, entropy, $H_k^i$, of a $k \times k$ block around each pixel, $p_i$, of original image, $f$, is used for classifying a pixel,  $p_i$, as a texture pixel, $p_t$, or as a cartoon pixel, $p_c$. According to calculated entropy values, pixel $p_i$ is preserved or replaced with another pixel, $p_a$, chosen from multi-scale ROF filtered images. $p_i$ and $p_a$ have the same spatial location $(x,y)$ . Here, multi-scale ROF filtered images denote images of which are results of  ROF Filtering applied to original image $f$ for different parameter, $\lambda$, values.

During my experiments with parameters of ROF model, I observed that at different scales TV Filter produces very nice edge preserving cartoon approximations of different parts of image. So, instead of using Gaussian filter to approximate cartoonized versions of a texture pixel $p_t$,  I used  TV Filter. For this study a hand tuned matching procedure is used for replacing a texture pixel $p_t$, with a cartoonized pixel $p_a$ which comes from an image selected from a multi-scale ROF Filtered images.

 The selection procedure is simple. I just applied a linear, bin based matching procedure. Multi-scale ROF Filtered images are treated as bins, and each pixel $p_i$ from original image $f$, is matched to a bin according to entropy value, $H_k^i$ , of a $k \times k$ block around it. So, if there are $5$ multi-scale ROF Filtered images, and $255$ entropy values, a pixel with entropy value $0$, is replaced with a pixel from an image filtered at scale $0$, namely the original image. A pixel with entropy value of $255$, is replaced with a pixel from an image filtered at scale $5$, namely the scale with smallest parameter $\lambda$, and with the smoothest image.

%
%
%
%
%

\subsection*{Output of Algorithm at 74 scales.}

Experiments on Barbara image is first made by choosing scale $74$ different scales. I applied ROF Filtering with $74$ different $\lambda$ values. $\lambda$ values starts from $0.005$, increase by $0.002$ at each step up to $0.159$. Two images with nearly same texture behaviour are given below, one of them is produced by my algorithm and other one is a result of only ROF Filtering. My algorithm achieves same texture elimination effect with better edge structure. Regions drawn in red are examples of edge behaviour to be observed.

 \newpage
\vspace*{50px}

\begin{figure}[H]
\centering
\subfigure[Image with my algorithm applied.]{%
\includegraphics[scale=0.3]{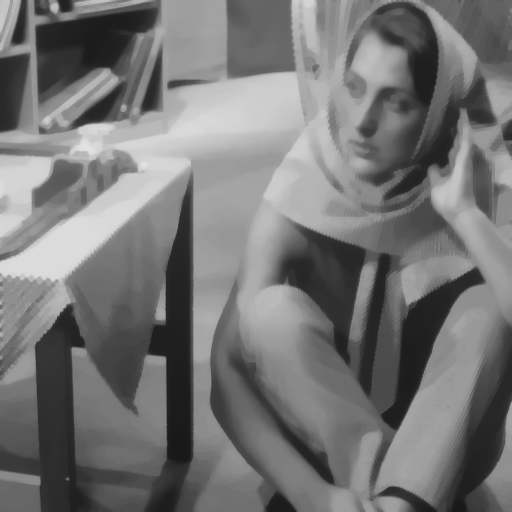}
\label{fig:subfigure1}}
\quad
\subfigure[Image with ROF. $\lambda = 0.025$]{%
\includegraphics[scale=0.3]{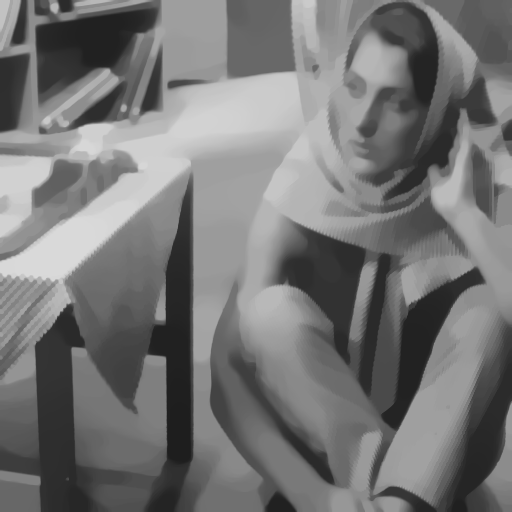}
\label{fig:subfigure2}}
\subfigure[Original Image]{%
\includegraphics[scale=0.4]{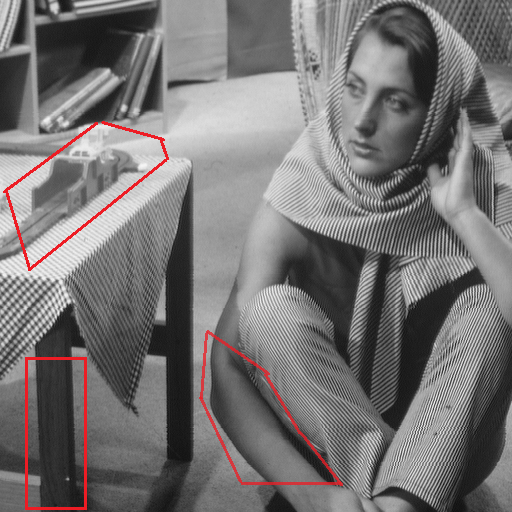}
\label{fig:subfigure3}}
\quad
\caption{A comparison of my algorithm and plain ROF.}
\label{fig:figure}
\end{figure}

%
%
%
%
%

\subsection*{Output of Algorithm at different scales.}

Finally, I compare results of my algorithm using different number of scales as bins. Also, result of a well known cartoon texture separation algorithm of Buades et al. is given as a reference for comparison. My algorithm is  new and selection of number of  scales  requires a through study.

%
%
%
%
%

\subsection*{Conclusions}
An entropy based cartoon texture separation algorithm is proposed and its results are evaluated. Although, algorithm is promising it requires detailed analysis and experiments with different block sizes of $k$, and scale parameters $\lambda$. Also, realistic entropy estimation procedures taking into account probabilistic dependency between pixels can be used to further develop proposed algorithm.

 \newpage
\vspace*{50px}

\begin{figure}[H]
\centering
\subfigure[Original Image]{%
\includegraphics[scale=0.4]{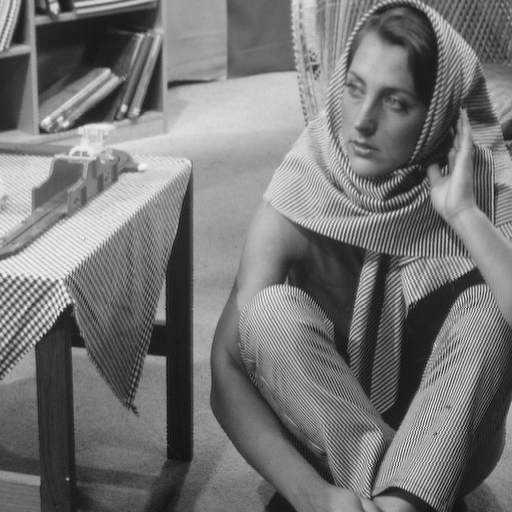}
\label{fig:subfigure1}}
\quad
\subfigure[Algorithm of Buades et al.]{%
\includegraphics[scale=0.4]{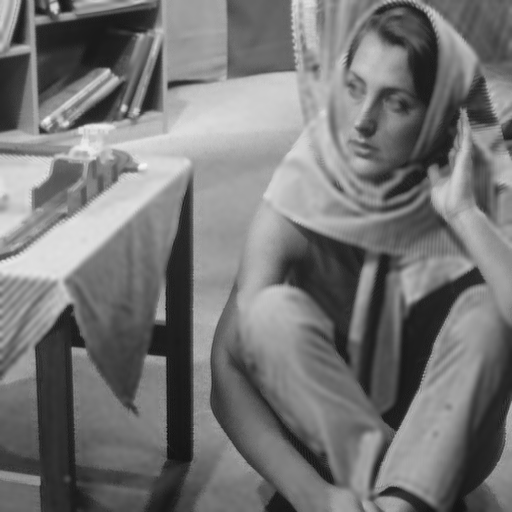}
\label{fig:subfigure2}}
\subfigure[My algorithm at $3$ ROF scales.]{%
\includegraphics[scale=0.4]{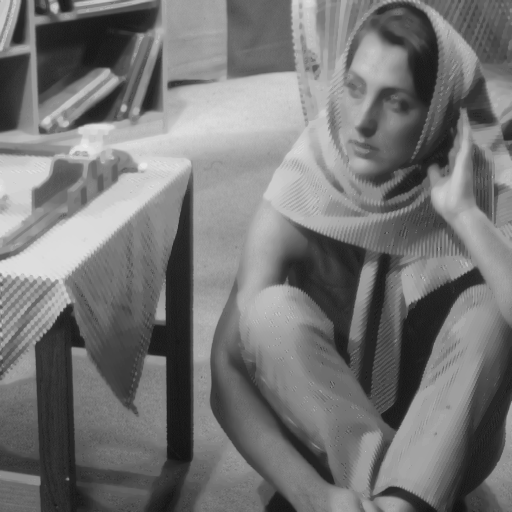}
\label{fig:subfigure3}}
\quad
\subfigure[My algorithm at $5$ ROF scales.]{%
\includegraphics[scale=0.4]{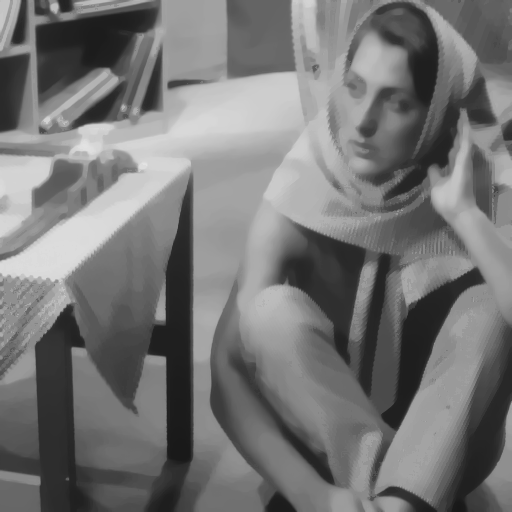}
\label{fig:subfigure4}}
\quad
\subfigure[My algorithm at $11$ ROF scales.]{%
\includegraphics[scale=0.4]{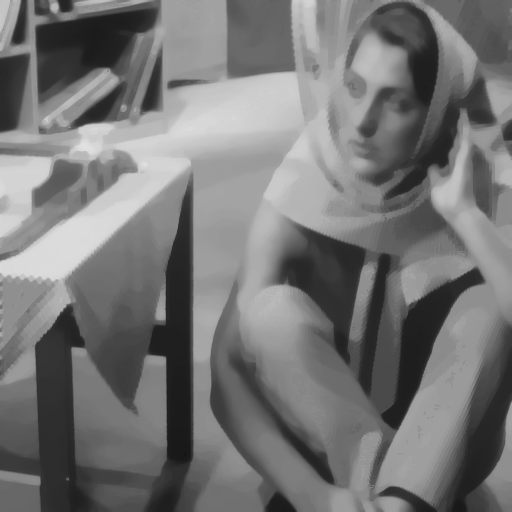}
\label{fig:subfigure5}}
\subfigure[My algorithm at $74$ ROF scales.]{%
\includegraphics[scale=0.4]{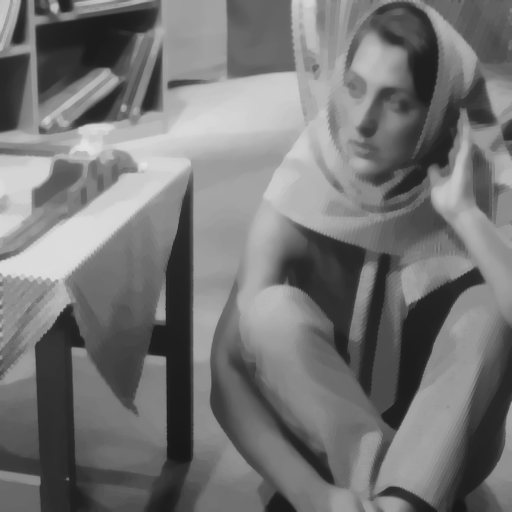}
\label{fig:subfigure6}}
\caption{Comparison of my algorithm at different scales with Buades et al.}
\label{fig:figure}
\end{figure}

%
%
%
%
%

\newpage

\vspace*{50px}

\subsection*{References}

\ \ \ \ \ [1] Rudin, Leonid I., Stanley Osher, and Emad Fatemi. "Nonlinear total variation based noise removal algorithms." Physica D: Nonlinear Phenomena 60.1 (1992): 259-268. \\

[2] Meyer, Yves. Oscillating patterns in image processing and nonlinear evolution equations: the fifteenth Dean Jacqueline B. Lewis memorial lectures. Vol. 22. AMS Bookstore, 2001. \\

[3] Yin, Wotao, Donald Goldfarb, and Stanley Osher. "A comparison of three total variation based texture extraction models." Journal of Visual Communication and Image Representation 18.3 (2007): 240-252. \\

[4] Zhu, Mingqiang, Stephen J. Wright, and Tony F. Chan. "Duality-based algorithms for total-variation-regularized image restoration." Computational Optimization and Applications 47.3 (2010): 377-400. \\

[5] Chan, Tony F., Gene H. Golub, and Pep Mulet. "A nonlinear primal-dual method for total variation-based image restoration." SIAM Journal on Scientific Computing 20.6 (1999): 1964-1977. \\

[6] http://pages.cs.wisc.edu/~swright/TVdenoising/ \\

[7] http://vivid.cse.psu.edu/texturedb/gallery/ \\

[8] Buades, Antoni, et al. "Fast cartoon+ texture image filters." Image Processing, IEEE Transactions on 19.8 (2010): 1978-1986. \\

%
%
%
%
%

\end{document}